\pdfoutput=1

\documentclass[11pt]{article}

\usepackage{acl}

\usepackage{threeparttable}

\usepackage{times}
\usepackage{latexsym}

\usepackage{booktabs}
\usepackage{subcaption} 

\usepackage{tabularx} 

\usepackage{caption}
\usepackage{geometry}
\usepackage{hyperref}
\usepackage{float}
\usepackage[T1]{fontenc}

\usepackage[utf8]{inputenc}

\usepackage{microtype}

\usepackage{inconsolata}

\usepackage{comment}
\usepackage{array}

\usepackage{graphicx} 

\usepackage{float} 

\usepackage{times}
\usepackage{latexsym}

\usepackage{listings}
\usepackage{color}
\usepackage{xcolor} 

\definecolor{codegreen}{rgb}{0,0.6,0}
\definecolor{codegray}{rgb}{0.5,0.5,0.5}
\definecolor{codepurple}{rgb}{0.58,0,0.82}
\definecolor{backcolour}{rgb}{0.95,0.95,0.92}

\usepackage{lipsum}

\definecolor{codegreen}{rgb}{0.58,0,0.8}
\definecolor{codegray}{rgb}{0.5,0.5,0.5}
\definecolor{darkred}{rgb}{139, 0, 0}
\definecolor{backcolour}{rgb}{0.95,0.95,0.92}

\lstdefinestyle{mystyle}{
    backgroundcolor=\color{backcolour},   
    commentstyle=\color{codegreen},
    keywordstyle=\color{blue},
    numberstyle=\tiny\color{codegray},
    stringstyle=\color{darkred},
    basicstyle=\ttfamily\footnotesize,
    breakatwhitespace=false,         
    breaklines=true,                 
    captionpos=b,                    
    keepspaces=true,                 
    numbers=left,                    
    numbersep=5pt,                  
    showspaces=false,                
    showstringspaces=false,
    showtabs=false,                  
    tabsize=2
}

\definecolor{greytext}{rgb}{0.5, 0.5, 0.5}

\lstdefinestyle{textstyle}{
    basicstyle=\color{greytext}\ttfamily\footnotesize,
    breakatwhitespace=false,         
    breaklines=true,                 
    captionpos=b,                    
    numbers=none, 
    showspaces=false,                
    showstringspaces=false,
    showtabs=false,                  
    tabsize=2
}

\newcommand\blfootnote[1]{%
  \begingroup
  \renewcommand\thefootnote{}\footnote{#1}%
  \addtocounter{footnote}{-1}%
  \endgroup
}

\title{Understanding the Effects of Noise in Text-to-SQL: \\ An Examination of the BIRD-Bench Benchmark} 

\author{Niklas Wretblad\textsuperscript{1,*} \quad
Fredrik Gordh Riseby\textsuperscript{1,*} \quad
Rahul Biswas\textsuperscript{2} \quad \\
\textbf{Amin Ahmadi\textsuperscript{2} \quad 
Oskar Holmström\textsuperscript{1}} \\
\textsuperscript{1}Linköping University, 
\textsuperscript{2}Silo AI \\
\texttt{niklas.wretblad@liu.se}
}

\begin{document}

\maketitle
\begin{abstract}
Text-to-SQL, which involves translating natural language into Structured Query Language (SQL), is crucial for enabling broad access to structured databases without  expert knowledge. However, designing models for such tasks is challenging due to numerous factors, including the presence of 'noise,' such as ambiguous questions and syntactical errors. This study provides an in-depth analysis of the distribution and types of noise in the widely used BIRD-Bench benchmark and the impact of noise on models. While BIRD-Bench was created to model dirty and noisy database values, it was not created to contain noise and errors in the questions and gold queries. We found that noise in questions and gold queries are prevalent in the dataset, with varying amounts across domains, and with an uneven distribution between noise types. The presence of incorrect gold SQL queries, which then generate incorrect gold answers, has a significant impact on the benchmark's reliability. Surprisingly, when evaluating models on corrected SQL queries, zero-shot baselines surpassed the performance of state-of-the-art prompting methods. We conclude that informative noise labels and reliable benchmarks are crucial to developing new Text-to-SQL methods that can handle varying types of noise. All datasets, annotations, and code are available at this \href{https://github.com/niklaswretblad/the-effects-of-noise-in-text-to-SQL}{URL}.
\end{abstract}

\section{Introduction}

Text-to-SQL\blfootnote{\textsuperscript{*}Equal Contribution} through large language models facilitates broader access to structured databases without requiring expert knowledge. To develop such models, high-quality open datasets and benchmarks are essential resources, and over the years, several benchmarks and datasets have been created. Early benchmarks, such as WikiSQL \cite{wikiSQL}, modeled simple scenarios, often with single-table queries, and following datasets attempts to closer approximate real-world scenarios: complex queries with join-statements over several tables \cite{spiderDataset}, unseen domain-specific datasets \cite{spiderDK} \cite{kaggleDBQA}, and noisy questions \cite{spiderSYN}. BIRD-Bench, a recent and challenging benchmark, aims to further close the gap between text-to-SQL research and real-world applications by for example containing large and dirty database values and requiring external knowledge \cite{birdbench}.

While BIRD-Bench does not explicitly introduce noise to the questions in the data, it could be that it is added inadvertently due to human error. For the same reason, noise is an essential aspect of real-world use cases, as human inputs often are ambiguous and contain syntactical errors. However, for the benchmark to be a helpful tool for judging model properties, such as noise handling, the data must be valid and inform us in what areas a model can be improved.

\begin{figure}[t]
  \centering
  \captionsetup{belowskip=-15pt}
  \includegraphics[width=\columnwidth]{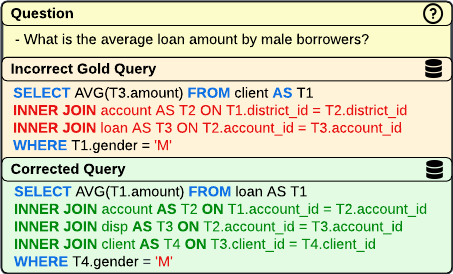}
  \caption{Example of an incorrect SQL query that generates the wrong gold reference answer for the given question. The JOIN operation incorrectly matches clients and accounts by district\_id. Due to the possibility of multiple clients and accounts in the same district, accounts are incorrectly associated with the wrong users.}
  \label{fig:sql_error}
\end{figure}

This paper continues the tradition of examining the suitability and limitations of open datasets and benchmarks. We specifically focus on how noise is represented in questions and queries in BIRD-Bench. We perform a qualitative analysis of what types of noise exist in the data and the noise distribution in specific domains. We then study the effects of noise on different models and prompting techniques, using both strong baselines and state-of-the-art methods.

We find that noise in questions and queries is prevalent, noise is unevenly distributed across domains, and categories of noise types are represented unequally in the data. Errors in gold SQL queries are also common and decrease the reliability of BIRD-Bench. When evaluating models on a dataset with corrected gold queries, the performance gap between zero-shot baselines and state-of-the-art prompting techniques is closed, questioning how we should interpret model performance on BIRD-Bench.

\section{Related Work}

\paragraph{Datasets} 

WikiSQL is a large Text-to-SQL dataset containing only simple SELECT and WHERE operations without nested queries or JOIN operations \cite{wikiSQL}. SPIDER \cite{spiderDataset} was later developed to approximate real-life scenarios more closely, requiring models to construct complex queries and understand the database schema. While complexity is a critical aspect of real use cases, variations of SPIDER have been created to contain noisy questions \cite{spiderSYN} and domain-specific questions \cite{spiderDK}.  BIRD-Bench was created to close the gap between academic research and real-world applications by introducing large and dirty database values, questions requiring external knowledge and optimizing SQL execution efficiency \cite{birdbench}.

\paragraph{Text-to-SQL Methods}

The notable gap in accuracy between automated systems (65.45\%) and human experts (92.96\%)\footnote{BIRD-Bench benchmark as of 2024-02-16 (https://bird-bench.github.io)}, highlights the need for ongoing developments in Text-to-SQL models.

Different approaches have been taken to create models capable of Text-to-SQL generation. 
A more traditional approach is to finetune LLMs on Text-to-SQL examples. While these models offer promising results, there is a performance gap to instruction-tuned LLMs, in particular GPT-4, that is adapted to the Text-to-SQL task through prompt engineering \cite{birdbench}. Prompts are often chained, where each prompt is applied to the task sub-problems, such as schema linking, decomposition of queries, and refinement of model generations \cite{dinsql, macSQL}. 

\paragraph{Noise in Datasets}
The contemporaneous work of \citet{macSQL} shows that ambiguous questions and incorrect SQL queries exist in BIRD-Bench. However, unlike our work, they do not study how noise varies across domains or how the identified noise and errors affect model performance.
\citet{surveyDLSQL} points out that database schemas often misalign with data entities, which may cause lexical or syntactic ambiguities affecting text-to-SQL models.

\begin{table*}[t]
\centering
\small
\begin{tabularx}{\textwidth}{Xcccccc}
\toprule
\textbf{Statistic} & \textbf{Financial} & \textbf{California Schools} & \textbf{Superhero} & \textbf{Toxicology} & \textbf{Thrombosis Prediction} \\
\midrule
Data points with noise & 52/106 (49\%) & 9/20 (45\%) & 3/20 (15\%) & 7/20 (35\%) & 8/20 (40\%) \\
Noisy questions & 44/106 (41.5\%) & 5/20 (25\%) & 2/20 (10\%) & 6/20 (30\%) & 3/20 (15\%) \\
Erroneous gold queries & 22/106 (20.7\%) & 8/20 (40\%) & 1/20 (5\%) & 2/20 (10\%) & 6/20 (30\%) \\
\bottomrule
\end{tabularx}
\caption{Statistics of the total amount of data points that contains errors and the amount of errors in questions and gold queries across five datasets. Note that a data point can have errors in both the question and SQL query.}
\label{tab:comprehensive_statistics}
\vspace{3mm}
\end{table*}

\begin{table*}[t]
\centering
\small
\begin{tabularx}{\textwidth}{Xcccccc}
\toprule
\textbf{Noise Type} & \textbf{Financial} & \textbf{California Schools} & \textbf{Superhero} & \textbf{Toxicology} & \textbf{Thrombosis Prediction} \\
\midrule
Spelling/Syntactical Errors & 23 & 2 & 1 & 4 & 2 \\
Vague/Ambiguous Questions & 17 & 1 & 1 & 1 & 1 \\
Incorrect SQL & 22 & 8 & 1 & 2 & 6 \\
Synonyms & 2 & 0 & 0 & 0 & 0 \\
String Capitalization & 7 & 0 & 0 & 0 & 0 \\
Question does not map to DB & 1 & 4 & 1 & 0 & 0 \\
\midrule
Total number of errors & 72 & 15 & 4 & 7 & 9 \\  
\bottomrule
\end{tabularx}
\caption{Distribution of different types of noise encountered in the domains.}
\label{tab:stats_errors}
\end{table*}

\section{Method}

\subsection{Data}
\label{data}

The BIRD-Bench dataset \cite{birdbench} is used in this paper as it is a recent and widely used dataset that is the most similar to real world scenarios. Since BIRD contains 12,751 samples across many domains and because of time-consuming human annotation, the main focus of the analysis is on the financial domain\footnote{This was also motivated by the fact this paper was a collaborative endeavor with the Swedish bank SEB.}, which includes queries related to banking operations.

The development set of the financial domain contains 106 question and SQL query pairs, which represent approximately 7.5\% of the data points in the development set and are structured around eight distinct tables, presented in full in Appendix \ref{app:financial_tables}. Each question is annotated with a difficulty level (simple, moderate, and challenging). The specific distribution is found in Figure \ref{fig:dist_accuracy}, in Appendix \ref{app:dist_acc}.

We selected four additional domains to validate our noise analysis of the financial domain and performed the same analysis on 20 randomly sampled questions from each domain. The selection was based on question difficulties and model accuracy of DIN-SQL\footnote{Results of DIN-SQL across domains were provided by the creators of DIN-SQL.} on each of the domains, as presented in Figure \ref{fig:dist_accuracy}. We selected \textit{California Schools} with low accuracy and simple questions, \textit{Superhero} with high accuracy and simple questions, \textit{Toxicology} with similar accuracy to the financial domain but more complex questions, and \textit{Thrombosis Prediction} with low accuracy and moderately difficult questions.

\subsection{Annotation of Noise}

All questions and SQL queries in the selected domains were annotated to determine whether they contained errors. The annotations were performed independently by two authors of this paper, fluent in English and experts in SQL. The identified errors were grouped based on similarity and named after the errors' common properties, as shown in Table \ref{tab:stats_errors}. The annotations were then used to generate two distinct datasets: one where SQL was corrected, and one where both SQL queries and noisy questions were corrected. 
We have published all annotations and cleaned datasets.\footnote{https://github.com/niklaswretblad/the-effects-of-noise-in-text-to-SQL}

\subsection{Models and Prompt Techniques}
\label{models-and-prompt-techniques}

Two models, GPT-3.5 and GPT-4, were used with three different prompting methods: zero-shot prompting as a baseline and the more advanced DIN-SQL \cite{dinsql} and MAC-SQL \cite{macSQL}. We used GPT-3.5 and GPT-4 for zero-shot prompting, but for the advanced prompting techniques, we only used GPT-3.5 since chaining prompts with GPT-4 was outside of the available resources for this project. We chose the models and prompting methods because they were the highest-performing publicly available models on BIRD-Bench at the time of writing.

Information about the database schema is crucial to generating correct queries for BIRD-Bench questions. DIN-SQL and MAC-SQL has a predefined format for adding the database schema. For the zero-shot model, we provide the database schema in-context in the form of SQL table creation statements, as this has been shown to improve accuracy compared to other formats \cite{promptDesignStrategies}. The prompt template for the zero-shot model is found in Appendix \ref{app:prompt} and the code for running experiments is publicly accessible.\footnotemark[\value{footnote}]

\begin{figure*}[t]
    \centering    
    \includegraphics[scale=0.4]{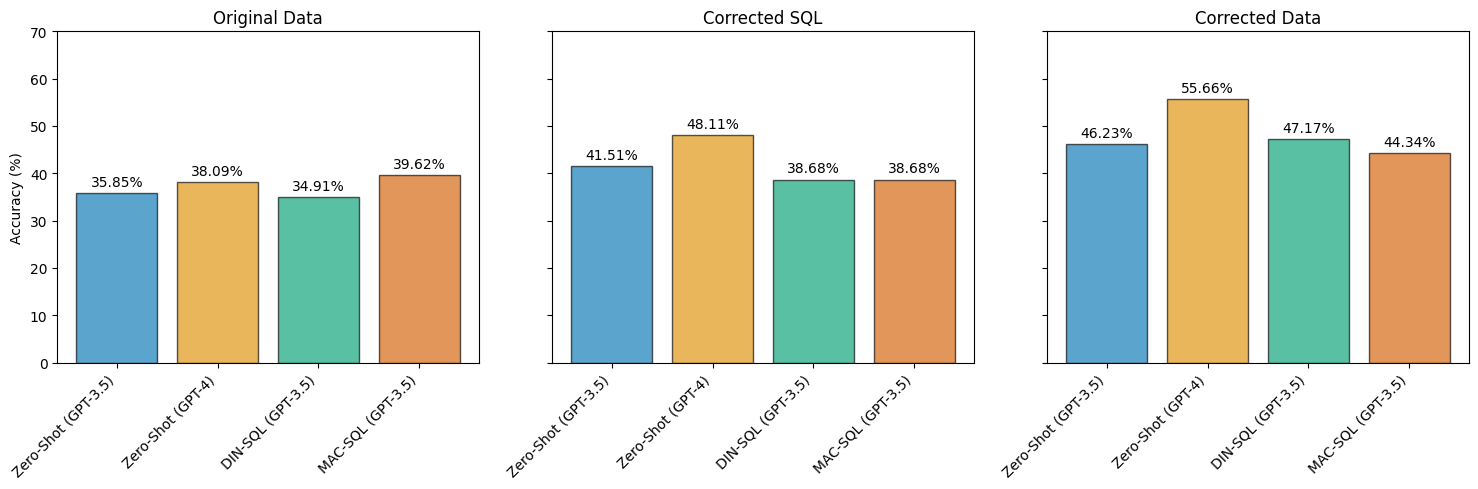} 
    \caption{Accuracy of various models on Bird-Bench's financial domain. Models are evaluated on the original data (left), corrected SQL queries (middle), and corrected SQL queries and corrected noisy questions.}
    \label{fig:execution_accuracy}
\end{figure*}

\section{Qualitative Analysis of Noise}

 Even though BIRD-Bench is not intended to represent noise in questions and SQL queries, our analysis reveals that noise exists in all studied domains to different extents. The financial domain exhibits the highest levels of noise at 49\% closely followed by the \textit{California\ Schools} domain at 45\%, as shown in Table \ref{tab:comprehensive_statistics}. In contrast, the \textit{Superhero} domain demonstrated the lowest noise levels, with only 15\% of data points containing errors. As presented in Section \ref{data} and Figure \ref{fig:dist_accuracy}, the \textit{Superhero} domain had the highest accuracy while having a similar distribution of question difficulties. This could indicate that model accuracy across tasks correlates with noise, which implies that noise in questions and SQL queries need to be carefully considered during dataset design.

The categories and absolute frequency of noise per dataset are presented in Table \ref{tab:stats_errors}, and both examples and descriptions of the noise types are presented in Appendix \ref{app:noise_examples}. Our analysis shows that spelling/syntactic errors and incorrect SQL queries were most prevalent in the financial domain. The presence of noise in questions is not necessarily undesirable, as it more closely mimics real-life scenarios. However, noise distribution across the categories is unequal. While this could be the real-world distribution, it might unfairly bias the benchmark towards models better at handling syntactical errors. Given the uneven distribution of errors and the lack of noise labels, the benchmark does not inform which noise types are challenging for current models and how models should improve.

A more severe issue is that all domains contained incorrect SQL queries, which are used for generating gold reference answers. An example of an erroneous SQL query is shown in Figure \ref{fig:sql_error}. These types of errors question the reliability of the benchmark to accurately determine model performance, which is explored in the next section.

\section{Impact of Noise on Model Performance}
\label{sec:model_impact}

We apply models to the original dataset, a dataset where SQL has been corrected, and a dataset where both SQL queries and noisy questions have been corrected. Figure \ref{fig:execution_accuracy} presents the results of a single evaluation for all models on all datasets.

MAC-SQL slightly outperforms DIN-SQL and the zero-shot baselines on the original dataset, where noise exists in both questions and queries. However, correcting SQL queries decreases MAC-SQL's performance, tying it with DIN-SQL as the poorest performers. Surprisingly, even the zero-shot GPT-3.5 baseline outperforms the more advanced DIN-SQL and MAC-SQL.
The dataset with corrected SQL queries could also be considered optimal since gold labels are correct and noise in questions is represented. Given the drastic re-ranking of models, it is relevant to question if BIRD-Bench is a reliable assessor of models and a useful tool to assist researchers in developing new methods for Text-to-SQL.

When evaluating models on the dataset with both questions and SQL queries corrected, the accuracy of all models increases significantly. While zero-shot GPT-4 performs the best, the remaining models perform similarly with DIN-SQL slightly ahead. Compared to the ideal scenario where only SQL queries are corrected, the presence of noise noticeably impacts all models' accuracy. However, models are not equally affected by noise as some models have a more pronounced increase in accuracy. This might suggests that the types of noise also affects models differently, which needs to be studied further.

\section{Conclusions and Future Work}

This paper analyzed the quality and distribution of noise in the BIRD-Bench benchmark for Text-to-SQL. We show that noise in both questions and SQL queries are prevalent, and noise is unevenly distributed across noise types and domains. Errors in gold SQL queries were common, decreasing the reliability of BIRD-Bench. Surprisingly, when evaluating models on corrected gold queries, zero-shot baselines surpassed more advanced prompting techniques. These findings highlight the necessity for developing benchmarks that can guide researchers in designing models that are more resistant to noise. Therefore, a significant improvement would be to label noise types across the dataset. In future work, we plan to study how large language models can be applied to noise classification, a new task that could also be critical in systems where Text-to-SQL is employed.

Overall, this study provides a deeper understanding of how noise is expressed in Text-to-SQL tasks and how noise and models interact, pinpointing areas for improvement in the BIRD-Bench dataset.

\newpage

\section*{Limitations}

While our study provides valuable insights regarding the influence of dataset noise in Text-to-SQL translation tasks, it has several limitations. As the analysis was performed mainly on the BIRD-Bench dataset's financial domain, our findings' generalizability may be limited. We only examined a small subset of other domains to validate our findings, which may represent only some of the noise distribution across domains.

Additionally, annotators may have introduced subjective bias during noise annotation, even though we attempt to minimize this by having two independent annotators. Further, our decision to categorize noise into six specific classes might have oversimplified the complexity and diversity of noise types in these benchmarks.

Our choice of models and prompting techniques could also be a potential limitation. We only employed two models, GPT-3.5 and GPT-4, and three different prompting methods. Evaluating a more comprehensive array of models and prompting techniques might have given a more comprehensive understanding of their performance under the influence of noise.

Lastly, the substantial effort required to correct SQL queries and noisy questions in the dataset may have introduced errors despite the review process. This might influence the model performances we report when evaluating models on the corrected datasets. 

\section*{Acknowledgments}
We extend our gratitude to Mohammadreza Pourreza for the results from the DIN-SQL model. We are also grateful to SEBx for their generous support and the provision of resources. Additionally, this research was partial funded by the National Graduate School of Computer Science in Sweden (CUGS).

\newpage
\bibliography{custom}

\appendix
\onecolumn

\newpage
\section{Appendix}
\label{sec:appendix}

\subsection{Database Schema of the Financial Domain}
\label{app:financial_tables}

\begin{figure*}[h]
    \centering    
    \includegraphics[scale=0.7]{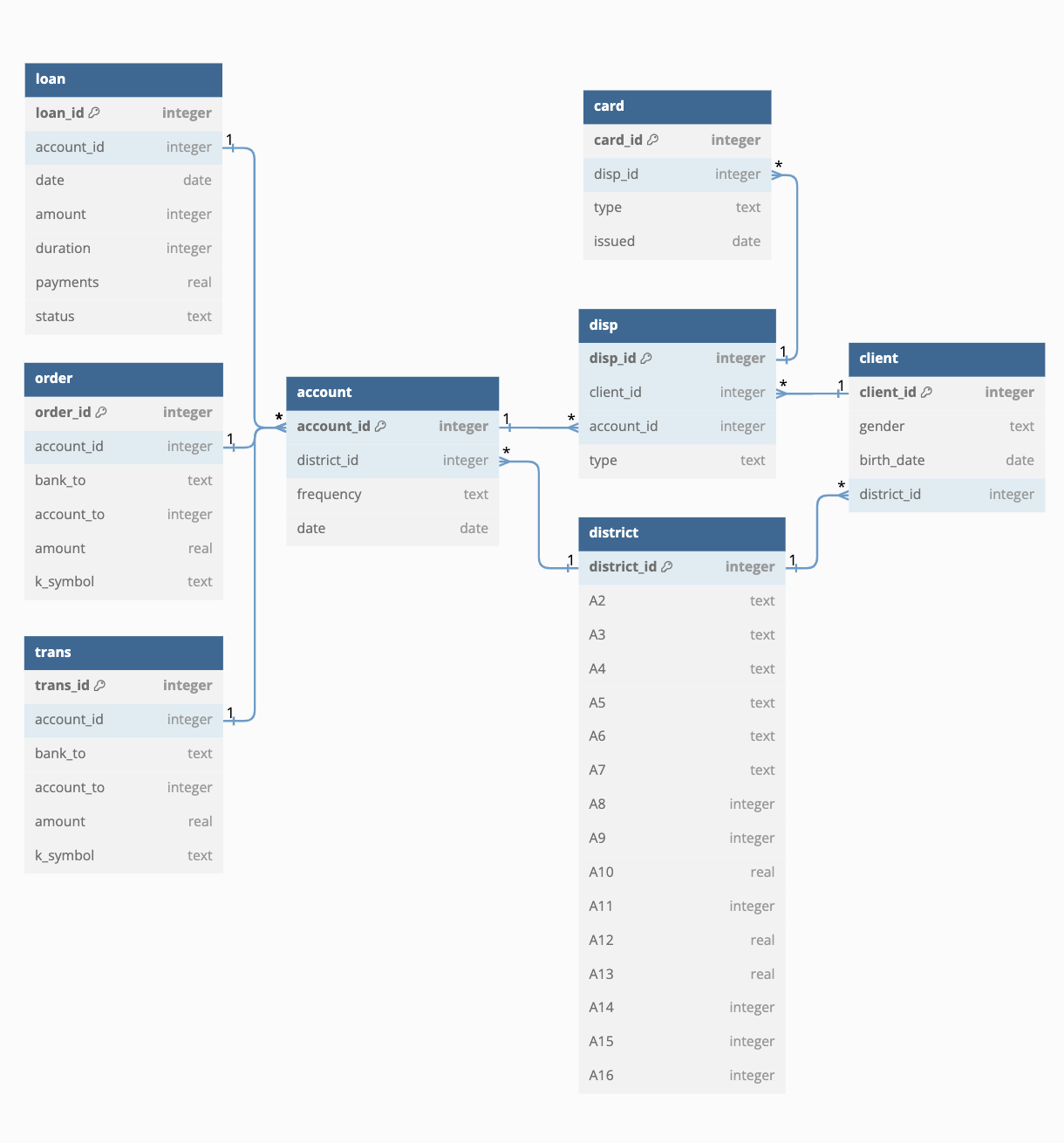} 
    \caption{Database schema of the database in the financial domain of BIRD-Bench.}
    \label{fig:database_schema}
\end{figure*}

\noindent Figure \ref{fig:database_schema} displays the database schema for the financial domain. This schema contains various tables, such as those for loans, transactions, accounts, cards and clients, all reflecting the financial orientation of the database. Descriptions of what information these tables contain are presented in Table \ref{tab:table_descriptions}. The database consists of 55 columns distributed across eight distinct tables. While the majority of the column names are intuitively understandable, some present interpretative challenges, as evident in the schema. An illustrative example is the district table, which incorporates 16 unique columns. This includes a column titled \textit{district\_ID} along with 15 other columns, ranging from \textit{A2} to \textit{A16}. The latter columns' names do not readily convey the nature of the data they hold, making them less intuitive to understand. In practice, a database schema will often be accompanied by a data dictionary or documentation that explains each table and column in detail. Such documentation would typically provide the context needed to fully understand the meaning of each element in the schema, the range of possible values for fields with unspecified types, and the business logic underlying the relationships. Without this additional documentation, fully interpreting and effectively using the database can be challenging as illustrated by the column names in the districts table. The BIRD-Bench dataset includes a unique feature for each question termed \textit{hint}. This feature is designed to offer insights or supplementary information corresponding to the specifics detailed in such database documentation. This feature is provided to all models described in \ref{models-and-prompt-techniques} for each question during the experiments. 

\begin{table}[h!]
\centering
\caption{Table descriptions of the tables in the database of the financial domain of BIRD-Bench.}
\begin{tabular}{@{}ll@{}}
\toprule
\textbf{Table Name} & \textbf{Description}                                      \\ \midrule
loan & Contains details of loans. \\
order & Holds information about monetary orders. \\
trans & Represents financial transactions. \\
account & Contains account information. \\
disp & Links clients to accounts (dispositions). \\
card & Contains details about cards issued. \\
client & Holds client information. \\
district & Contains details about districts or regions. \\ \bottomrule
\end{tabular}
\label{tab:table_descriptions}
\end{table}

Further, the lines in Figure \ref{fig:database_schema} between the tables represent relationships, where the nature of the relationship is indicated by the shape of the tail end of the lines where they connect to each table. 
A one-to-many relationship is indicated by the line beginning with a single line and the one digit above it, and then ending in a crow's foot (three lines) at the opposite end. For example, an account can have multiple orders, transactions, dispositions, and loans associated with it, but each of those entities is only linked to one account. An account can have many loans, but one loan is exclusively only linked to one account, which makes sense. Further, clients and accounts are related through the disposition table in a many-to-many relationship. An account can have many different clients associated with it, for example, one client listed as the owner of the account and multiple other clients listed as users for the account. This could for example be practical for sharing an account in a family, where one parent could be the owner of the account and then multiple other family members listed as users. A single client can also be related to many different accounts in the other way around.

\subsection{Distribution of question difficulty and model accuracy}
\label{app:dist_acc}

\begin{figure*}[h]
    \centering    
    \includegraphics[scale=0.37]{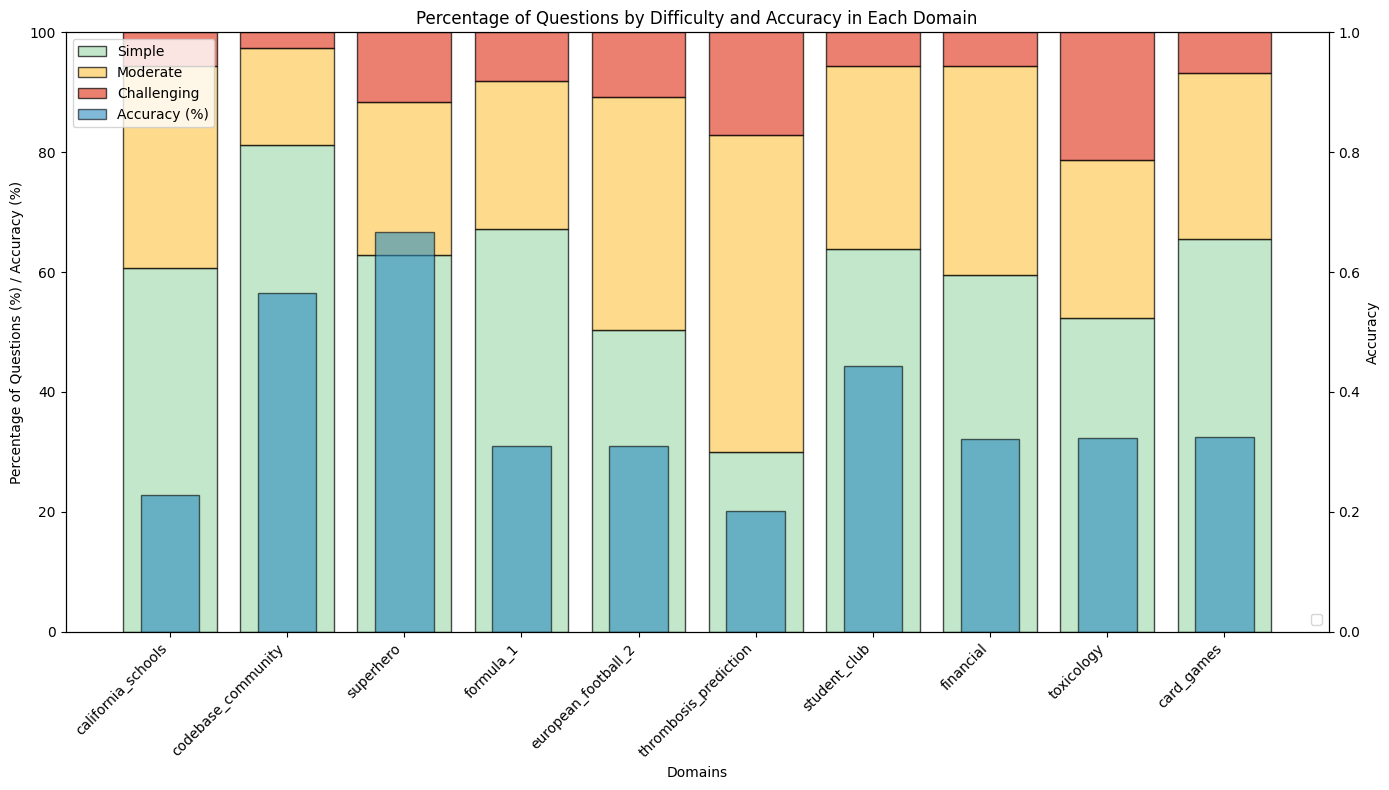} 
    \caption{Distribution of question difficulties and execution accuracy of the DIN-SQL model on the different domains of the BIRD-Bench development set.}
    \label{fig:dist_accuracy}
\end{figure*}

\subsection{Prompt Templates}
\label{app:prompt}

\begin{lstlisting}[language=Python, caption={Zero-Shot Prompting Template.}, label=zero_shot_template, frame=single, framesep=3mm, xleftmargin=21pt, tabsize=4]
"""Database schema in the form of CREATE_TABLE statements:

{database_schema}

Using valid SQL, answer the following question based on the 
tables provided above. 

Hint helps you to write the correct sqlite SQL query.
Question: {question}
Hint: {evidence}
DO NOT return anything else except the SQL query."""
\end{lstlisting}

\noindent The prompt template underlying the zero-shot models described in Section \ref{models-and-prompt-techniques} can be found in Listing \ref{zero_shot_template}. The prompt integrates a given question, the associated database schema, an instruction directing the LLM to generate valid SQL, and a hint provided by the BIRD-Bench dataset. The hint is designed to offer insights or supplementary
information needed in order to accurately interpret the database schema and to correctly convert the question into a SQL query. Note that the other models implemented in this research is also provided with this feature.

\newpage

\subsection{Examples of Errors and Corrections}
\label{app:noise_examples}

This section provides examples of erroneous data points and their corrections from the different error categories found in Table \ref{tab:comprehensive_statistics}.

\subsection*{\textbf{Example 1: Spelling/Syntactical Error}}
\begin{figure*}[h]
    \centering    
    \includegraphics[scale=1]{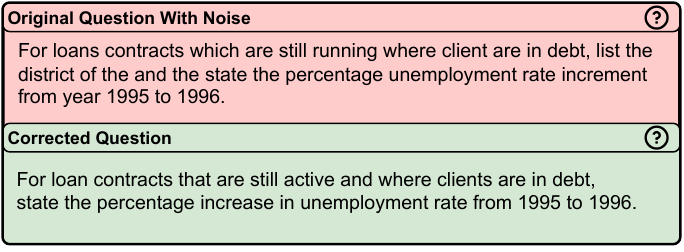} 
    \caption{Question with ID 125 from the development set of BIRD-Bench which contains syntactical errors and a corrected version of the question. }
    \label{fig:semantic_error}
\end{figure*}

In Figure \ref{fig:semantic_error}, an example question with a syntactical error is provided, representing the question with ID 125 from the financial domain in the BIRD-Bench development set. The grammatical structure of the question complicates the interpretation of its meaning for a human reader and makes it difficult to understand which information it is asking for. Therefore, there is a chance that an LLM might also misinterpret the question. A corrected version of the question can be seen in the figure.

\subsection*{\textbf{Example 2: Ambiguous/Vague Question}}

\begin{figure*}[h]
    \centering    
    \includegraphics[scale=1]{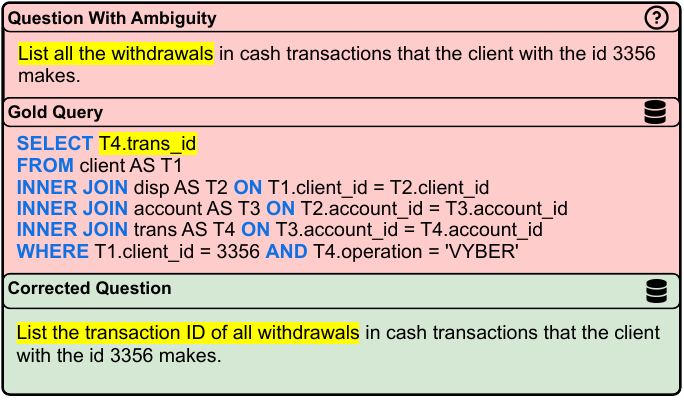} 
    \caption{Question, gold SQL query and a corrected version of the question corresponding to the data point with ID 159 from the development set of BIRD-Bench, showcasing an error in the ambiguous/vague category. }
    \label{fig:ambiguous_error}
\end{figure*}

Figure \ref{fig:ambiguous_error} displays the data point with ID 159 from the financial domain of the development set of BIRD-Bench. It contains an error which were grouped into the ambiguous/vague question category. The challenge lies in the natural language question's ambiguity, specifically in the phrase \textit{``List all the withdrawals...''} This ambiguity revolves around determining which columns to return when executing the SQL query.

\newpage

\subsection*{\textbf{Example 3: Incorrect Gold SQL}}

\begin{figure}[h]
  \centering
  \includegraphics[scale=1.3]{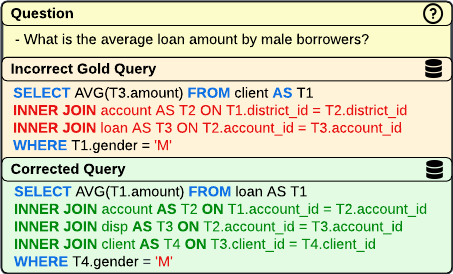}
  \caption{Example of an incorrect SQL query that generates the wrong gold reference answer for the given question. The JOIN operation incorrectly matches clients and accounts by district\_id. Due to the possibility of multiple clients and accounts in the same district, accounts are incorrectly associated with the wrong users.}
  \label{fig:sql_error2}
\end{figure}

Figure \ref{fig:sql_error2} showcases an incorrect golden SQL query found in the data point with ID 132 of the financial domain of the development set of BIRD-Bench. The JOIN operation incorrectly matches clients and accounts by district\_id. Due to the possibility of multiple clients and accounts in the same district, accounts are incorrectly associated with the wrong users.

\newpage
\subsection*{Example 4: Synonyms}
\begin{figure*}[h]
    \centering    
    \includegraphics[scale=1]{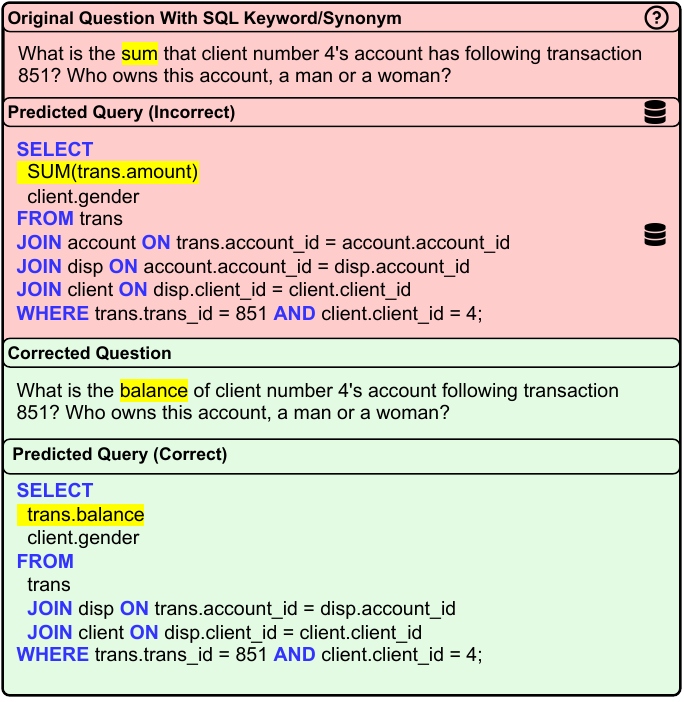} 
    \caption{Question from data point with ID 177 from the development set of BIRD-Bench containing a difficult synonym, a corrected version of the question with the synonym replaced and corresponding predicted SQL queries by the DIN-SQL (GPT-3.5) model described in Section \ref{models-and-prompt-techniques}. Showcases the difficulty of synonyms on model predictions.}
    \label{fig:synonym-error}
\end{figure*}

Figure \ref{fig:synonym-error} demonstrates how specific wordings can complicate interpretation for an LLM. The term \textit{sum}, being both a SQL keyword and a descriptor, led to the LLM's literal interpretation and the incorrect summation of a transaction. The actual intent was to inquire about the transaction's balance or amount. A rephrased question resulted in the LLM generating the correct SQL query, fetching the intended information, as seen in the figure. 

\newpage
\subsection*{\textbf{Example 5: String Capitalization}}
\begin{figure*}[h]
    \centering    
    \includegraphics[scale=1]{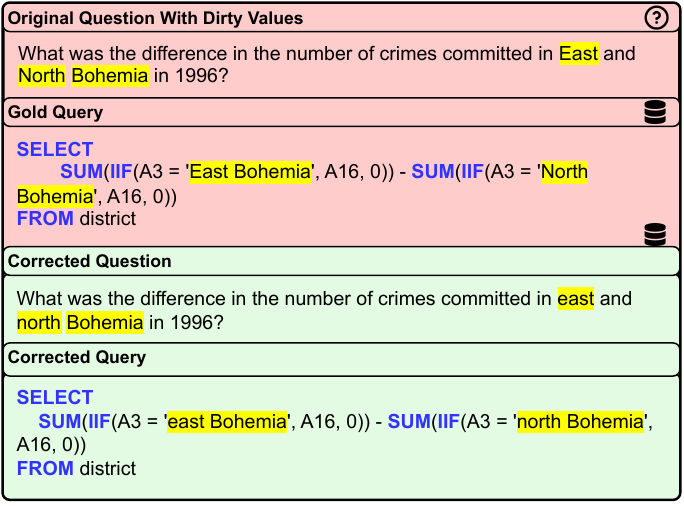} 
    \caption{Example Ambiguous.}
    \label{fig:string-capitalization-error}
\end{figure*}

As a consequence of SQL being a case-sensitive language when comparing string values in a query, the way a question is formulated regarding the use of uppercase or lowercase letters when asking for a specific value affects the result. This is because the LLM will most likely use the specific entry as given when generating the query, unless it has knowledge of the case used for different entries in the database. Therefore, in Figure \ref{fig:string-capitalization-error}, an example is provided where the terms "East" and "North" are mentioned with initial capital letters, as is commonly the case. However, the entries for these column values are in lowercase in the database, which means the question needs to account for this for the LLM to be able to generate a correct query. The corrected question and the SQL query generated from it can also be seen in Figure \ref{fig:string-capitalization-error}.

\subsection*{Example 6: Database Schema Non-Alignment}

\begin{table}[H]
\caption{Examples of questions that does not map to the database schema and accompanying descriptions of why they do not.}
\label{tab:query_feasibility}
\centering
\begin{tabular}{@{}p{0.5\textwidth}p{0.4\textwidth}@{}}
\toprule
\textbf{Incorrect Question} & \textbf{Description} \\ \midrule
What is the disposition ID of the client who made \$5100 USD transaction on 1998/9/2? & The question asks for a single disposition ID, which does not reflect that there is a one-to-many relation between client and disposition, and most likely it won't be possible to return a single ID. \\ \midrule
List out the account numbers of clients who are youngest and have highest average salary? & There is no information about salaries of specific clients in the database. \\ 
\bottomrule
\end{tabular}
\end{table}

Table \ref{tab:query_feasibility} showcases two questions improperly aligned with the database schema's structure, meaning the questions cannot be accurately answered with data from the database. The structure of the database schema can be viewed in Figure \ref{fig:database_schema}. The first question overlooks the possibility of clients having multiple dispositions, as well as multiple disposition IDs. The second question instead cannot be converted into a SQL correctly since there is no information in the database about salaries of specific clients.

\end{document}